\documentclass[sn-mathphys-num]{sn-jnl}
\usepackage{graphicx}%
\usepackage{multirow}%
\usepackage{amsmath,amssymb,amsfonts}%
\usepackage{amsthm}%
\usepackage{mathrsfs}%
\usepackage[title]{appendix}%
\usepackage{xcolor}%
\usepackage{textcomp}%
\usepackage{manyfoot}%
\usepackage{booktabs}%
\usepackage{algorithm}%
\usepackage{algorithmicx}%
\usepackage{algpseudocode}%
\usepackage{listings}%

\usepackage{anyfontsize}
\usepackage{ragged2e} 
\usepackage{multirow}
\usepackage{array}

\usepackage{pgffor}

\DeclareMathOperator*{\argmin}{argmin}

\begin{document}

\title[Interval Regression: A Comparative Study with Proposed Models]{Interval Regression: A Comparative Study with Proposed Models}

\author*[1]{\fnm{Tung} \sur{L Nguyen}}\email{tln229@nau.edu}
\author[2]{\fnm{Toby} \sur{Dylan Hocking}}\email{toby.dylan.hocking@usherbrooke.ca}

\affil*[1]{\orgdiv{School of Informatics, Computing, and Cyber Systems}, \orgname{Northern Arizona University}, \orgaddress{\street{S San Francisco}, \city{Flagstaff}, \postcode{86011}, \state{Arizona}, \country{USA}}}
\affil[2]{\orgdiv{Département d'informatique}, \orgname{Université de Sherbrooke}, \orgaddress{\street{Sherbrooke QC J1K 2R1}, \city{Quebec}, \country{Canada}}}

\abstract{Regression models are common for a wide range of real-world applications.
However, for the data collection in practice, target values are not always precisely known; instead, they may be represented as intervals of acceptable values.
This challenge has led to the development of Interval Regression models. 
In this study, we provide a comprehensive review of existing Interval Regression models and propose alternative models for comparative analysis. 
Experiments are conducted on both real-world and synthetic datasets to offer a broad perspective on model performance. 
The results demonstrate that no single model is universally optimal, highlighting the importance of selecting the most suitable model for each specific scenario.}


\maketitle

\section{Introduction} \label{sec:intro}
Regression models are widely used in real-world applications. 
However, due to some reasons (errors in measurement, data collection procedures or human factors), the target value is \textit{not} a single value but rather an interval of acceptable values.
One example is in survival analysis \citep{example_medical_survival, example_medical}, the exact survival time of a patient is often uncertain, though it is typically known that the patient has survived for at least a certain number of years.
In economic forecasting \citep{example_econ}, and environmental studies \citep{example_environmental}, precise measurements at a specific time period are unreliable, leading researchers to use the confidence interval of measurements over time.
Similarly, in engineering \citep{example_engineering}, multiple measurements of a single statistic are taken, and the confidence interval is considered as the target. 
These uncertainties motivate the study of Interval Regression.

\begin{table}[t]
\centering
\begin{tabular}{l|l|l|l}
\cmidrule(lr){2-4}
                    & \multicolumn{2}{c|}{training dataset}                                                 &                          \\ \cmidrule(lr){2-3}
                    & features (predictors)         & target (ground truth)                       & prediction (response)    \\ \midrule
Standard Regression & $\mathbf{x} \in \mathbf{R}^m$ & $y \in \mathbf{R}$                          & $\hat{y} \in \mathbf{R}$ \\ \midrule
Interval Regression & $\mathbf{x} \in \mathbf{R}^m$ & $(y_l, y_u) \in \mathbf{R}^2, y_l \leq y_u$ & $\hat{y} \in \mathbf{R}$
\\ \bottomrule
\end{tabular}
\caption{Comparison of Standard Regression versus Interval Regression setting.}
\label{tab:interval_vs_standard}
\end{table}

The term ``regression" typically refers to Standard Regression (or Point Regression), where each instance is associated with a single target value.
This study focuses on Interval Regression setting, where each instance is associated with an interval of acceptable values, called \textit{target interval}.
There are 4 types of target intervals $(y_l, y_u)$ based on the end values: uncensored ($-\infty < y_l = y_u < \infty$), right-censored ($-\infty < y_l < y_u = \infty$), left-censored ($-\infty = y_l < y_u < \infty$), and interval-censored ($-\infty < y_l < y_u < \infty$). 
Although many studies are titled with ``Interval Regression", such as \citep{interval_regression_analysis_1, interval_regression_analysis_2, interval_regression_analysis_3, interval_regression_analysis_4}, the setting of these studies differs from this work since they focus on predicting an interval for each instance. 
Instead, in this study's Interval Regression setting, the model predicts a single value for each instance, see Table \ref{tab:interval_vs_standard}.

\begin{figure}[!tb]
\includegraphics[width=1\textwidth]{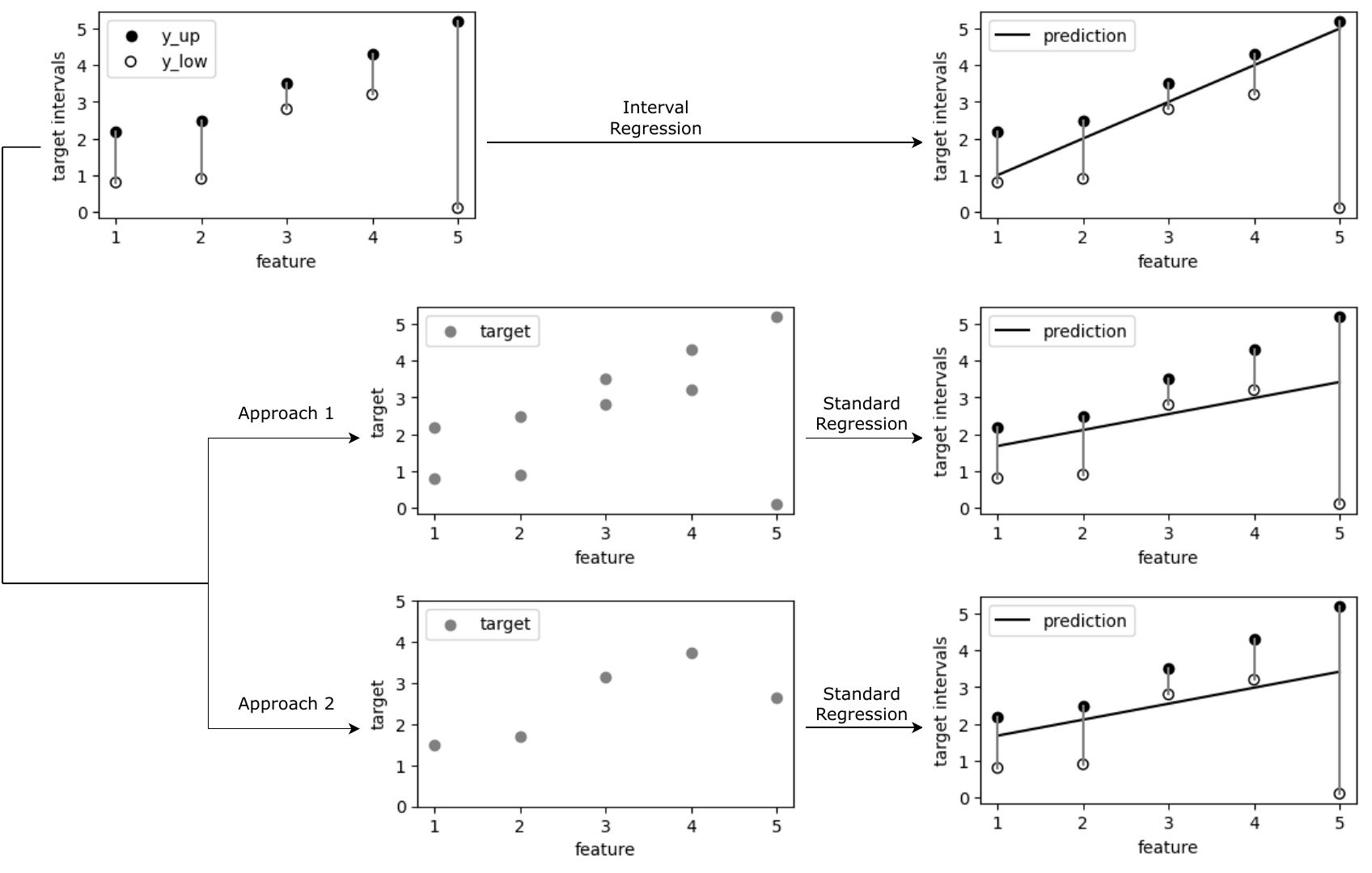}
\caption{Example of converting Interval Regression into Standard Regression. In approach 1, each interval instance is represented by two endpoints, while in approach 2, it is represented by the midpoint. The goal of Interval Regression is to predict a value falls within the target interval. This example shows that these conversion approaches perform poorly in Interval Regression setting, so they are not recommended.}
\label{fig:convert_interval_to_standard}
\end{figure}

In some studies, Interval Regression models are trained by converting the problem into Standard Regression: instead of using intervals as targets, the target interval is represented by discrete points, which are then treated as Standard Regression targets.
Using the same notations in Table \ref{tab:interval_vs_standard} for $\mathbf{x}$ as the features set and $(y_l, y_u)$ as the target interval,
\citet{mmit} transform the instance $\big(\mathbf{x}, (y_l, y_u)\big)$ into two points, \((\mathbf{x}, y_l)\) and \((\mathbf{x}, y_u)\), called Interval-CART model whereas \citet{rit} convert it into its midpoint, \((\mathbf{x}, \frac{y_l + y_u}{2})\) -- Approach 1 and 2 in Figure \ref{fig:convert_interval_to_standard}.
While these approaches are reasonable, it overlooks the interval information by representing intervals by finite values and fails to handle left-censored or right-censored intervals, resulting in wasted data.
So the performance of these models is worse compared to the models specifically designed for interval targets, see the results in \citep{mmit, rit} and Figure \ref{fig:convert_interval_to_standard} example.


To date, existing Interval Regression models for this setting include Linear \citep{rigaill2013penalty, rit, opart_pen_mlp25}, Tree-based \citep{mmit, aft_xgboost} and Accelerated Failure Time (AFT) models family \citep{aft_semi, aft, aft_huang2006regularized, aft_cai2009regularized}.
Details on these models can be found in Section \ref{sec:models}.
\\
\\
\noindent\textbf{Contribution.} This study proposes some alternative Interval Regression models and offers a comprehensive performance comparison of all models (existing and proposed) across a variety of datasets, including real-world and synthetic.

\section{Interval Regression Models} \label{sec:models}
This section details Interval Regression models, including existing and proposed ones. 
See Table \ref{tab:models} for all of the employed models in this study.
The notation used hereafter follows Table \ref{tab:interval_vs_standard}, where $\mathbf{x} \in \mathbf{R}^m$ is the model input (a vector of $m$ features), $(y_l, y_u) \in \mathbf{R}^2$ where $y_l \leq y_u$ is the target interval, and $\hat{y} \in \mathbf{R}$ is the model prediction.

\subsection{Existing models}

\paragraph{Max Margin Interval Regression (linear)} This linear model was introduced by \citet{rigaill2013penalty}, and its extended theoretical foundations were later studied by \citet{rit}.
The prediction formula is given by:
\begin{equation} \label{eq:linear_prediction}
    \hat{y} = \mathbf{x} \cdot \beta + \beta_0
\end{equation}
where $\beta \in \mathbf{R}^m$ and $\beta_0 \in \mathbf{R}$ are the parameters. To estimate these parameters, the authors extended the Standard Regression loss function — the Squared Error (SE) — into a generalized form called the \textit{Hinge Squared Error}.
The error value between prediction $\hat{y}$ and target interval $(y_l, y_u)$ is defined as: 

\begin{table}[!t] 
\begin{tabular}{l|l|l|l} 
\toprule
\textbf{model name} & \textbf{source code} & \textbf{programming language} & \textbf{citation} \\ \midrule
linear                 & \citet{penaltyLearning} & R                                      & \citet{rigaill2013penalty}             \\ 
mmit                   & \citet{mmit_package} & Python binding C++                     & \citet{mmit}                           \\ 
aft\_xgb               & \citet{xgb_package} & Python                                 & \citet{aft_xgboost}                    \\ \midrule
constant                    & \href{https://github.com/lamtung16/ML_IntervalRegression}{paper GitHub repo} & Python binding C++                                 & additional baseline                               \\ \midrule
knn                    & \href{https://github.com/lamtung16/ML_IntervalRegression}{paper GitHub repo} & Python                                 & proposed                               \\ 
mlp                    &\href{https://github.com/lamtung16/ML_IntervalRegression}{paper GitHub repo}& Python                                 & proposed                               \\ 
mmif                   &\href{https://github.com/lamtung16/ML_IntervalRegression}{paper GitHub repo}& Python binding C++                     & proposed                               \\ 
\bottomrule
\end{tabular}
\caption{List of the employed models in this study with corresponding source.}
\label{tab:models}
\end{table}

$$l\Big(\hat{y}, (y_l, y_u)\Big) = 
\begin{cases} 
    (y_l - \hat{y} + \epsilon)^2, & \mbox{if } \hat{y} < y_l + \epsilon \\
    0, & \mbox{if } y_l + \epsilon \leq \hat{y} \leq y_u - \epsilon \\
    (\hat{y} - y_u + \epsilon)^2, & \mbox{if } \hat{y} > y_u - \epsilon
\end{cases}$$

\noindent where $\epsilon \geq 0$ is the margin length.
This error becomes SE when $y_l +\epsilon = y_u - \epsilon$.

\citet{rit} proposed a model called \textit{Regression with Interval Targets} (RIT), which is essentially the same as model \eqref{eq:linear_prediction}, but uses \textit{Hinge Absolute Error} instead:

$$l\Big(\hat{y}, (y_l, y_u)\Big) = 
\begin{cases} 
    |y_l - \hat{y} + \epsilon|, & \mbox{if } \hat{y} < y_l + \epsilon \\
    0, & \mbox{if } y_l + \epsilon \leq \hat{y} \leq y_u - \epsilon \\
    |\hat{y} - y_u + \epsilon|, & \mbox{if } \hat{y} > y_u - \epsilon
\end{cases}$$

\begin{figure}[t]
\includegraphics[width=1\textwidth]{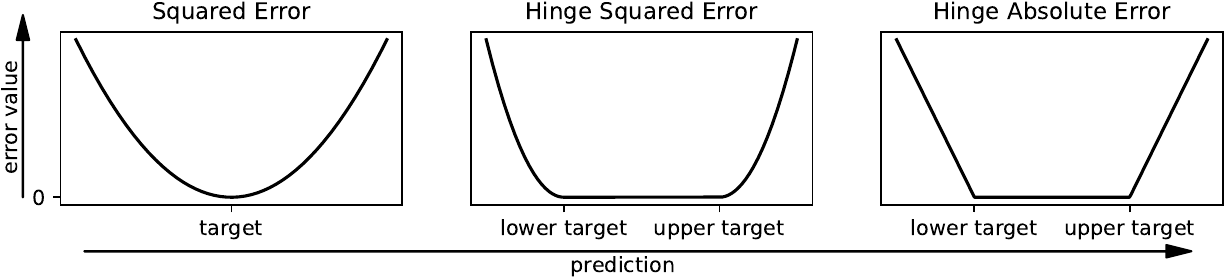}
\caption{Visualization of the Loss Functions: Error values relative to predictions and targets.}
\label{fig:loss_func}
\end{figure}

\noindent The general Interval Regression loss function can be written in this short form:

\begin{equation} \label{eq:lossfunction}
    l\Big(\hat{y}, (y_l, y_u)\Big) = \Big( \text{ReLU}(y_l - \hat{y} + \epsilon)\Big)^p + \Big(\text{ReLU}(\hat{y} - y_u + \epsilon) \Big)^p
\end{equation}

\noindent where $p \in \{1, 2\}$. See Figure \ref{fig:loss_func} for the visual comparison between hinge errors and SE.

To estimate the parameters $\beta$ and $\beta_0$ in \eqref{eq:linear_prediction}, apply a gradient descent algorithm with respect to $\beta$ and $\beta_0$ to minimize the sum of Hinge Error \eqref{eq:lossfunction} over the training set.
\citet{rigaill2013penalty} employed the Fast Iterative Shrinkage-Thresholding Algorithm (FISTA) \citep{fista}, while \citet{rit} employed stochastic gradient descent (SGD).

\paragraph{Maximum Margin Interval Trees (MMIT)} This Tree model was introduced by \citet{mmit} as an alternative nonlinear model to \textit{Max Margin Interval Regression}.
The prediction is given by:
\[
\hat{y} = \mathcal{T}(\mathbf{x})
\]
where \(\mathcal{T}\) represents the tree model.
The tree has the same structure as CART (Classification And Regression Tree) \citep{cart}. 
The only difference lies in the regression value for each leaf which is a set of $S$ instances $\{\mathbf{x}^i, (y_l^i,y_u^i)\}_{i=1}^S$: instead of taking the mean of all targets like CART does, it chooses a constant value $c$ that minimizes the mean Hinge Error \eqref{eq:lossfunction} between this constant and the target intervals:
\begin{equation*}
    \hat{y} = \argmin_c \frac{1}{S} \sum_{i=1}^{S} l\big(c, (y_l^i, y_u^i)\big), \quad c \in \mathbf{R}
\end{equation*}
\noindent A heuristic algorithm is used to speed up the estimation of the mean value for each leaf by considering a finite set of candidates, rather than solving for the optimal $c$:

\begin{equation}
    \hat{y} = \argmin_c \frac{1}{S} \sum_{i=1}^{S} l\big(c, (y_l^i, y_u^i)\big), \quad c \in \{y_l^i, y_u^i, \frac{y_l^i + y_u^i}{2}|y_l^i > -\infty, y_u^i < \infty\}_{i = 1}^S
\end{equation}

\noindent The training procedure follows the same operations as in CART to minimize the sum of Hinge Error \eqref{eq:lossfunction} over the training set.

\paragraph{AFT Model in XGBoost (aft\_xgb)}
The AFT model in XGBoost, introduced by \citet{aft_xgboost}, is an ensemble tree-based model designed for survival analysis.
The model makes predictions as follows:  
\[
y_p = \mathbf{T}(\mathbf{x}),
\]
where $\mathbf{T}$ represents the ensemble of trees in XGBoost.
This model can handle only non-negative target intervals.
To extend its applicability to real-valued target intervals $(y_l, y_u) \in \mathbf{R}^2$, \citet{aft_xgboost} apply an exponential transformation to the original interval.
After making a prediction $y_p$ in the exponential space, the result is mapped back to the original scale using the logarithm transformation:
$$\hat{y} = \log y_p = \log \mathbf{T}(\mathbf{x})$$
Training this model is based on minimizing the negative log-likelihood function (loss function), which incorporates all training instances:
\begin{equation} \label{eq:aft_loss}
l_{AFT}\Big(\hat{y}, (y_l, y_u)\Big)=
    \begin{cases}
      -\log \big(f_Z(\frac{\log y_u - \hat{y}}{\sigma})\times \frac{1}{\sigma y_u}\big) &\text{if $y_l = y_u$}\\
      -\log \Big[ F_Z\big( \frac{\log y_u - \hat{y}}{\sigma} \big) - F_Z\big(\frac{\log y_l - \hat{y}}{\sigma}\big) \Big] &\text{if $y_l < y_u$}
    \end{cases}
\end{equation}
\noindent where $\sigma > 0$ is the scale parameter.
The Probability Density Function $f_Z(z)$ and Cumulative Distribution Function $F_Z(z)$ depends on the chosen distribution:
\begin{itemize}
    \item normal: $f_Z(z) = \frac{\exp (-z^2/2)}{\sqrt{2\pi}}$, \phantom{00}  $F_Z(z) = \frac{1}{2}\big( 1 + \text{erf}(\frac{z}{\sqrt{2}}) \big)$
    \item logistic: $f_Z(z) = \frac{e^z}{(1+e^z)^2}$, \phantom{0000} $F_Z(z) = \frac{e^z}{1+e^z}$
    \item extreme: $f_Z(z) = e^ze^{-\exp z}$, \phantom{0} $F_Z(z) = 1 - e^{-\exp z}$
\end{itemize}

\begin{figure}[t]
\includegraphics[width=1\textwidth]{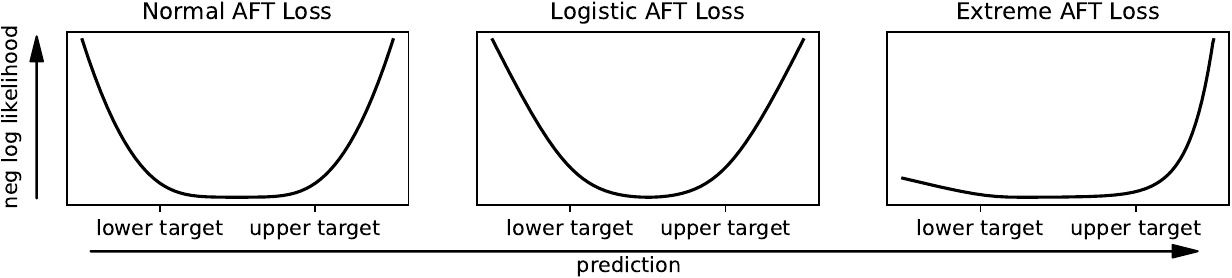}
\caption{Visualization of the AFT XGBoost Loss (negative log-likelihood) for 3 distributions.}
\label{fig:aft_loss_func}
\end{figure}

Figure \ref{fig:aft_loss_func} visualizes the loss function for each distributions, which can be compared to Figure \ref{fig:loss_func} to observe their similarities. 
The model is trained using the gradient boosting algorithm \citep{gradient_boosting} to minimize the loss value \eqref{eq:aft_loss} over the training dataset given the distribution and the scale parameter $\sigma$.

\subsection{Proposed Models}
In addition to existing models, this study proposes several alternative models, including machine learning architectures such as KNN, Neural Networks and Forests, to provide a more comprehensive comparison.
\paragraph{K-Nearest Neighbors (KNN)} KNN is a classical model for Standard Regression, which motivates its consideration for Interval Regression.
The operation follows the same steps as the standard KNN regression model \citep{knn}, with the only difference being how the regression value is determined from the \( k \) nearest neighbors.
We propose utilizing the MMIT regression function. 
Specifically, we treat the set of \( k \) nearest neighbors as a single leaf in a MMIT, where the regression value is a constant that minimizes the mean Hinge Error \eqref{eq:lossfunction} between this constant and the \( k \) target intervals.
The prediction $\hat{y}$ based on the target intervals of the \( k \) nearest neighbors, denoted as \( \{(y_l^i, y_u^i)\}_{i=1}^k \), is formulated as:
\[
    \hat{y} = \argmin_c \frac{1}{k} \sum_{i=1}^{k} l\big(c, (y_l^i, y_u^i)\big), \quad c \in \{y_l^i, y_u^i, \frac{y_l^i + y_u^i}{2}|y_l^i > -\infty, y_u^i < \infty\}_{i = 1}^k
\]

\paragraph{Multilayer Perceptron (MLP)} Since MLP is a commonly used neural network in Standard Regression models \citep{uni_approximator}, its application in Interval Regression is worth considering.
The training process of this model is the same as in the linear model \textit{Max Margin Interval Regression}.

\paragraph{Maximum Margin Interval Forest (MMIF)} The idea of using ensembles of MMITs was mentioned in Chapter 6 (Discussion and Conclusions) of \citep{mmit}. 
Since MMIT is a tree for the Interval Regression setting, its extension to construct a Forest is straightforward.  
The operation of MMIF follows the same principles as the standard Random Forest regressor \citep{rf}, with the key distinction being that MMITs are used instead of CART.

\begin{table}[!t]
\begin{tabular}{l|l|l|c|c|c|c}
\multicolumn{1}{c|}{\multirow{2}{*}{\textbf{dataset}}} & \multicolumn{1}{c|}{\multirow{2}{*}{\textbf{features}}} & \multicolumn{1}{c|}{\multirow{2}{*}{\textbf{instances}}} & \multicolumn{4}{c}{\textbf{features pattern/quality}} \\
\multicolumn{1}{c|}{}                         & \multicolumn{1}{c|}{}                          & \multicolumn{1}{c|}{}                           & linear   & nonlinear  & noisy  & excessive  \\ \midrule
auto93                                       & 62                                            & 82                                             &$\checkmark$       &            &        &$\checkmark$         \\
autohorse                                    & 59                                            & 159                                            &$\checkmark$       &            &        &$\checkmark$         \\
autompg                                      & 25                                            & 392                                            &$\checkmark$       &            &        &            \\
autoprice                                    & 15                                            & 159                                            &          &$\checkmark$         &        &            \\
baskball                                     & 4                                             & 96                                             &$\checkmark$       &            &        &            \\
bodyfat                                      & 14                                            & 252                                            &$\checkmark$       &            &        &            \\
breasttumor                                  & 40                                            & 286                                            &$\checkmark$       &            &        &            \\
cholesterol                                  & 26                                            & 299                                            &$\checkmark$       &            &$\checkmark$     &            \\
cleveland                                    & 26                                            & 299                                            &$\checkmark$       &            &        &            \\
cloud                                        & 10                                            & 108                                            &          &$\checkmark$         &        &            \\
cpu                                          & 36                                            & 209                                            &          &$\checkmark$         &        &            \\
echomonths                                   & 13                                            & 106                                            &$\checkmark$       &            &        &            \\
elusage                                      & 13                                            & 55                                             &$\checkmark$       &            &        &            \\
fishcatch                                    & 15                                            & 158                                            &$\checkmark$       &            &        &            \\
fruitfly                                     & 8                                             & 125                                            &$\checkmark$       &            &$\checkmark$     &            \\
housing                                      & 14                                            & 506                                            &          &$\checkmark$         &        &            \\
lowbwt                                       & 23                                            & 189                                            &          &$\checkmark$         &        &            \\
machine.cpu                                  & 6                                             & 209                                            &          &$\checkmark$         &        &            \\
mbagrade                                     & 3                                             & 61                                             &          &$\checkmark$         &$\checkmark$     &            \\
meta                                         & 54                                            & 264                                            &$\checkmark$       &            &        &$\checkmark$         \\
pbc                                          & 29                                            & 276                                            &$\checkmark$       &            &        &            \\
pollution                                    & 15                                            & 60                                             &          &$\checkmark$         &        &            \\
pwlinear                                     & 10                                            & 200                                            &          &$\checkmark$         &        &            \\
pyrim                                        & 27                                            & 74                                             &          &$\checkmark$         &$\checkmark$     &            \\
sensory                                      & 36                                            & 576                                            &$\checkmark$       &            &$\checkmark$     &            \\
servo                                        & 19                                            & 167                                            &          &$\checkmark$         &        &            \\
sleep                                        & 7                                             & 51                                             &$\checkmark$       &            &        &            \\
stock                                        & 9                                             & 950                                            &          &$\checkmark$         &        &            \\
strike                                       & 23                                            & 625                                            &$\checkmark$       &            &        &            \\
triazines                                    & 60                                            & 186                                            &          &$\checkmark$         &        &$\checkmark$         \\
veteran                                      & 13                                            & 137                                            &$\checkmark$       &            &        &            \\
vineyard                                     & 3                                             & 52                                             &$\checkmark$       &            &        &            \\
wisconsin                                    & 32                                            & 194                                            &          &$\checkmark$         &        &            \\ \midrule
simulated.abs                                & 20                                            & 200                                            &          &$\checkmark$         &$\checkmark$     &            \\
simulated.linear                             & 20                                            & 200                                            &$\checkmark$       &            &$\checkmark$     &            \\
simulated.sin                                & 20                                            & 200                                            &          &$\checkmark$         &$\checkmark$     &            \\ \midrule
pharynx                                      & 218                                           & 195                                            &          &$\checkmark$         &        &$\checkmark$         \\
lymphoma.mkatayama                               & 258                                           & 41                                             &          &$\checkmark$         &$\checkmark$     &$\checkmark$         \\
lymphoma.tdh                                 & 258                                           & 23                                             &$\checkmark$       &            &$\checkmark$     &$\checkmark$        
\\\bottomrule
\end{tabular}
\caption{All datasets in this study. Linear means the pattern between features and targets is likely linear, while nonlinear indicates otherwise. Noisy refers to cases where more than half of the features are irrelevant, excessive describes the dataset having more features than instances.}
\label{tab:datasets}
\end{table}

\begin{figure}[t]
\vskip 0.2in
\begin{center}
\centerline{\includegraphics[width=1\textwidth]{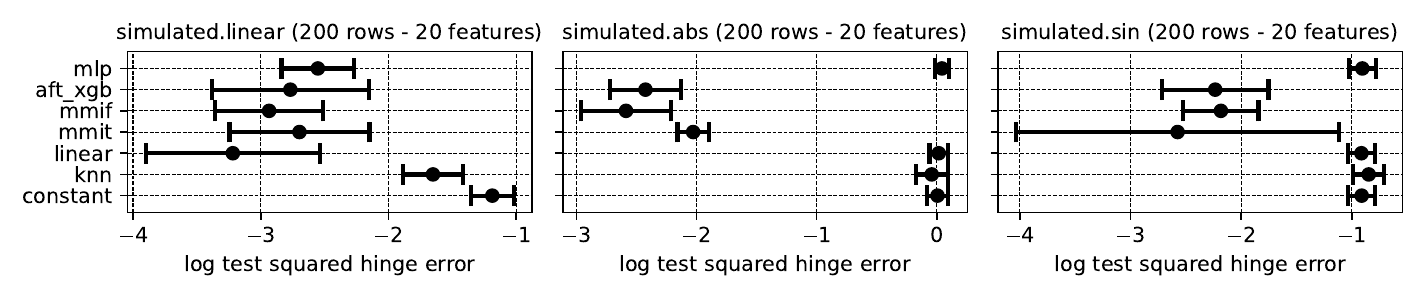}}
\caption{The mean and standard deviation of the log of test squared hinge errors from simulated datasets. The Linear model performs best when the dataset is linear. In nonlinear datasets, Tree-based models achieve the best performance.}
\label{fig:err_simulated}
\end{center}
\vskip -0.2in
\end{figure}

\section{Experiments} \label{sec:experiments}
\paragraph{Experimental Datasets}  
To ensure a comprehensive comparison across different scenarios, we evaluate all the models on a diverse set of datasets. 
Specifically, we use 36 real-world datasets from the UCI repository \citep{uci} including 3 biology datasets having excessive features, along with 3 simulated datasets - identical to those in \citep{mmit}.
Each simulated dataset consists of 200 instances with 20 features, where 19 are noise and one is the true feature \( x \). 
The target intervals are defined based on the linear function of $x$, \( \sin(x) \), or the absolute value of \( x \).
The real-valued outputs in these data sets were transformed into censored intervals, a small vertical shift was added by sampling a value from $\mathcal{N}(0, \frac{y_l}{10})$ and adding it to both interval bounds.
All datasets are described in Table \ref{tab:datasets} and are accessible here: \href{https://github.com/lamtung16/ML_IntervalRegression/tree/main/data}{study datasets link}.

\paragraph{Evaluation Metrics}
Each dataset is divided into 5 similar sized folds. 
Each fold is used as the test set, while the remaining 4 folds are combined to form the train set.
For each train/test pair, the model is trained on the train set and used to make predictions on the test set. 
The test set is denoted as having \( M \) instances with corresponding target intervals \( \{(y_l^i, y_u^i)\}_{i=1}^M \), and the predictions set is \( \{\hat{y}^i\}_{i=1}^M \). 
The Mean Squared Hinge Error is given by:
\begin{equation} \label{eq:eval_loss}
    \frac{1}{M} \sum_{i=1}^{M} \left( \text{ReLU}(y_l^i - \hat{y}^i) \right)^2 + \left( \text{ReLU}(\hat{y}^i - y_u^i) \right)^2
\end{equation}
which is equivalent to using Hinge Error \eqref{eq:lossfunction} with $p=2$ and $\epsilon=0$. Each train/test pair produces one error values. 
From these 5 error values, the mean and standard deviation are computed.  
Two metrics are then considered: \textbf{performance} and \textbf{consistency}. 
A smaller mean error indicates better performance, while a smaller standard deviation indicates better consistency.

\begin{figure}[t]
\vskip 0.2in
\begin{center}
\centerline{\includegraphics[width=1\textwidth]{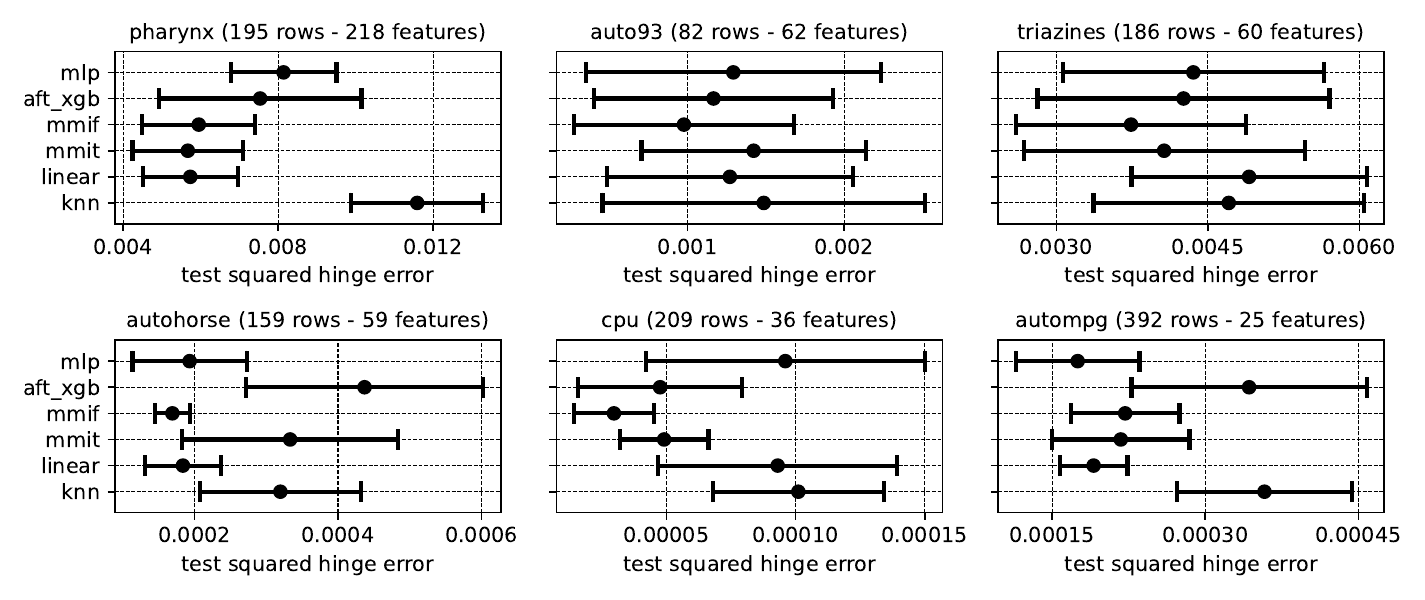}}
\caption{The mean and standard deviation of test squared hinge errors for datasets with a high number of features. Tree-based models generally perform well due to their inherent feature selection mechanism. On the other hand, while MLP with ReLU activation function is a more generalized Linear model, it fails to outperform the Linear model. One reason for this is that when a dataset contains a majority of noisy features, MLP cannot effectively reduce their impact on predictions in the same way that a Linear model with L1 regularization can.}
\label{fig:err_real_1}
\end{center}
\vskip -0.2in
\end{figure}

\subsection{Model Configuration}
Except for the AFT model in XGBoost, the Squared Hinge Error is used as the loss function for both the cross-validation process and model training.
The configuration for each model below is applied to a single train/test set pair.

\paragraph{Constant (Featureless)} 
This model predicts a single value from the train set using only target intervals.
The value is a constant that minimizes the Mean Squared Hinge Error \eqref{eq:eval_loss} between this constant and the set of target intervals.

\paragraph{Linear} 
The model \textit{Max Margin Interval Regression} model is implemented with L1 regularization.
The regularization parameter starts at 0.001, geometrically increasing by a factor of 1.2 until no features remain, with cross-validation (\texttt{cv} = 5) on the train set to determine the optimal L1 value.

\paragraph{MMIT}  
A cross-validation (\texttt{cv} = 5) was used to select the optimal hyperparameters, including \texttt{max\_depth} \{0, 1, 5, 10, 20, \(\infty\)\} and \texttt{min\_sample} \{0, 1, 2, 4, 8, 16, 20\}.

\paragraph{AFT model in XGBoost}
Cross-validation (\texttt{cv} = 5) was performed to select the optimal hyperparameters, following the same grid search described in \citet{aft_xgboost}:
\begin{itemize}
    \item \texttt{learning\_rate}: 0.001, 0.01, 0.1, 1.0
    \item \texttt{max\_depth}: 2, 3, 4, 5, 6, 7, 8, 9, 10
    \item \texttt{min\_child\_weight}: 0.001, 0.1, 1.0, 10.0, 100.0
    \item \texttt{reg\_alpha}: 0.001, 0.01, 0.1, 1.0, 10.0, 100.0
    \item \texttt{reg\_lambda}: 0.001, 0.01, 0.1, 1.0, 10.0, 100.0
    \item \texttt{aft\_loss\_distribution\_scale}: 0.5, 0.8, 1.1, 1.4, 1.7, 2.0
\end{itemize}

\begin{figure}[t]
\vskip 0.2in
\begin{center}
\centerline{\includegraphics[width=1\textwidth]{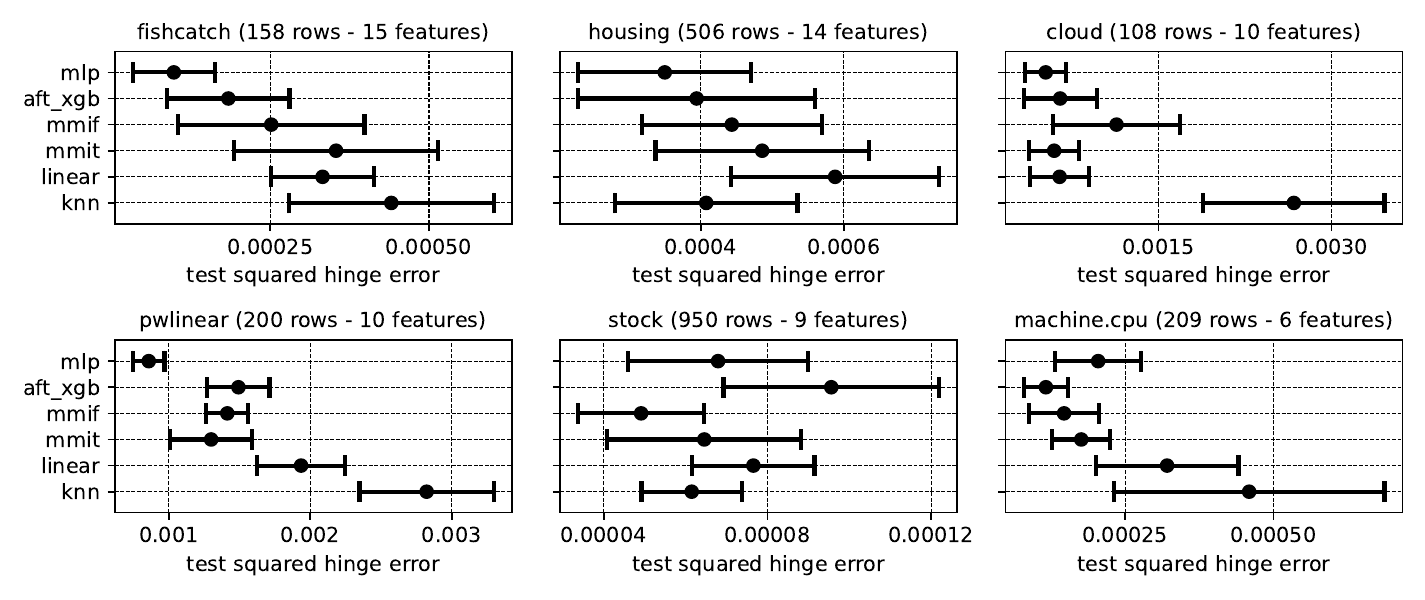}}
\caption{The mean and standard deviation of test squared hinge errors for datasets with a moderate number of features. In these scenarios, MLP generally performs well.}
\label{fig:err_real_2}
\end{center}
\vskip -0.2in
\end{figure}

\paragraph{KNN}
The Euclidean distance metric is applied to normalized features (mean 0, standard deviation 1) to determine the nearest neighbors. 
A cross-validation (\texttt{cv} = 5) is used to select the optimal value of \( K \).
The candidate values for \( K \) range from 1 to \( \lceil\sqrt{n}\rceil \), where \( n \) is the number of train instances.  

\paragraph{MLP} A cross-validation (\texttt{cv} = 5) is used to determine the optimal hyperparameters:  
\begin{itemize} 
    \item \texttt{num\_layer}: \{1, 2\}  
    \item \texttt{hidden\_layer\_size}: \{5, 10, 20\}  
    \item \texttt{activation}: \{ReLU, Sigmoid\}  
\end{itemize}
The model is trained using the Adam optimizer with a fixed learning rate of 0.001. 

\paragraph{MMIF}  
MMIF is an ensemble of 100 MMITs.
For each MMIT, two-thirds of the train dataset with one-third of the original features are randomly selected for training, while the remaining instances serves as the out-of-bag (OOB) set for validation. 
A cross-validation (\texttt{cv} = 5) is used to select the optimal hyperparameters for each MMIT including \texttt{max\_depth} \{2, 5, 10, 15, 20, 25\} and \texttt{min\_split\_samples} \{2, 5, 10, 20, 50\}.
Each MMIT has its own OOB error, which is used to determine the aggregation rule.
Let \(\mathcal{T}_i\) be an MMIT with OOB error \(e_i\). 
The weight assigned to \(\mathcal{T}_i\) is given by
$w_i = \frac{\frac{1}{e_i}}{\sum_{j=1}^{100} \frac{1}{e_j}}$
and the final prediction from MMIF is computed as $\hat{y} = \sum_{i=1}^{100} w_i \mathcal{T}_i(\mathbf{x})$,
where $\mathbf{x}$ is the set of features. 

\begin{table}[t]
\centering
\begin{tabular}{l|rrrrrrr|rrrrrrr}
\toprule
 & \multicolumn{7}{c|}{Performance} & \multicolumn{7}{c}{Consistency} \\
\midrule
 & 1st & 2nd & 3rd & 4th & 5th & 6th & 7th & 1st & 2nd & 3rd & 4th & 5th & 6th & 7th \\
\midrule
constant & 2 & 2 & 3 & 1 & 4 & 7 & 20 & 4 & 0 & 2 & 3 & 3 & 8 & 19 \\
knn & 4 & 4 & 2 & 6 & 6 & 16 & 1 & 4 & 3 & 3 & 7 & 7 & 12 & 3 \\
linear & 3 & 11 & 6 & 8 & 5 & 5 & 1 & 7 & 7 & 8 & 7 & 7 & 3 & 0 \\
mmit & 2 & 10 & 9 & 3 & 9 & 4 & 2 & 2 & 13 & 4 & 8 & 8 & 2 & 2 \\
mmif & 15 & 5 & 8 & 7 & 4 & 0 & 0 & 12 & 9 & 9 & 4 & 3 & 2 & 0 \\
aft\_xgb & 4 & 4 & 3 & 4 & 7 & 5 & 12 & 4 & 1 & 7 & 2 & 3 & 10 & 12 \\
mlp & 9 & 3 & 8 & 10 & 4 & 2 & 3 & 6 & 6 & 6 & 8 & 8 & 2 & 3 \\
\bottomrule
\end{tabular}
\caption{The comparison of the performance and consistency of 7 models across 39 datasets. For example, the constant model has the worst performance (rank 7th) in 20 out of 39 datasets, model mmif has the best consistency (rank 1st) in 12 out of 39 datasets.}
\label{tab:comparison}
\end{table}

\subsection{Results}
For each of the 39 datasets, 7 models were implemented, with each model producing 5 hinge error values.
Figure \ref{fig:err_simulated} shows the comparison on 3 highly noisy simulated datasets, where 95\% of the features are noise (19 out of 20).
Figure \ref{fig:err_real_1} shows the comparison on 6 slightly unstable datasets, where the feature-to-instance ratio is high (from 0.06 to 1.12).
Figure \ref{fig:err_real_2} shows the comparison on 6 fairly stable datasets, where the feature-to-instance ratio is low (from 0.009 to 0.09).
For more detailed comparison of all 39 datasets—including comparisons on hinge loss, confidence intervals, and IQR boxes—see the supplemental material here: \ref{sec:supplemental}.

\section{Discussions and Conclusions} \label{sec:discussion}
\paragraph{Dataset Variations}
To ensure a comprehensive comparison, the quality of the datasets is important.
The datasets employed have a variety of characteristics, including high- and low-dimensional data, real-world and synthetic sources, linear and nonlinear patterns, as well as low- and high-noise conditions. 
An overview of the dataset qualities is provided in Table \ref{tab:datasets}.

\paragraph{Comprehensive Comparison}
The performance and consistency rankings are in Table \ref{tab:comparison}, 1st means the best and 7th means the worst. 
About the performance, it's clear that MMIF outperforms other models in most cases. 
Linear, MMIT and MLP are comparable, while Constant, KNN, and the AFT model in XGBoost show the poorest performance.
About the consistency, MMIF stands out as the most consistent. 
This is expected, as MMIF being an ensemble model, relies on a group of MMITs to make the final prediction, which helps improve its stability.
Again, Constant, KNN, and the AFT model in XGBoost show the lowest consistency.

\paragraph{No single model is optimal for all scenarios}  
As shown in Table \ref{tab:comparison}, there is no universally optimal model for every scenario.
However, if complexity is not a primary concern, and the goal is to select the model with the best overall performance and consistency, MMIF is likely the first choice.  
Alternatively, if model complexity and training time are critical factors, MMIT is preferable, as it's less complex than MMIF, require significantly fewer training resources, and still maintain reliable performance.

\paragraph{MMIT vs MMIF} 
This is a comparison worth considering.
MMIF was proposed with the expectation that an ensemble of MMITs would outperform a single MMIT, and the results confirm this hypothesis. 
Specifically, across 39 datasets, MMIF outperformed MMIT in 28 of them. 
This makes sense: MMIF, being an ensemble of MMITs, is able to reduce overfitting by not relying on a single MMIT.

\paragraph{Discussion on the AFT Model in XGBoost and potential future work}
The AFT model in XGBoost is the complex models in this study, but its performance and consistency do not justify the complexity. 
It has a limitation in that it can only handle non-negative target intervals. 
To make the model applicable to a general Interval Regression setting, the target intervals must be transformed using exponential function. 
This introduces a weakness, as the model becomes highly sensitive when the dataset contains many left-censored target intervals. 
When predictions are made on the exponential scale, if the predicted value is close to zero, the mapping back to the original scale can result in large negative values, as seen in some test cases where predictions like $-10^6$ were made.
\[
(-\infty, y_u) \overset{\text{exp}}{\longrightarrow} (0, \exp y_u) \overset{\text{predict}}{\longrightarrow} y_p \approx 0 \overset{\text{log}}{\longrightarrow} \hat{y} \approx -\infty \overset{\text{evaluate}}{\longrightarrow} \text{Big test error}
\]
The experiments conducted in \citep{aft_xgboost} (Figure 2.b) indicate that the AFT model in XGBoost performs among the worst across their six simulated datasets when compared to the Linear and MMIT. 

A potential direction for future work is to transform left-censored intervals into interval-censored intervals by replacing all lower bounds of \(-\infty\) with a finite value. This adjustment aims to reduce the model's sensitivity to left-censored intervals and mitigate the tendency to produce excessively large negative predictions.

\paragraph{Code Available} All experiment code is available in this GitHub repository: \url{https://github.com/lamtung16/ML_IntervalRegression} for reproducibility and further exploration of the models used in this study.

\clearpage
\bibliography{ref}

\clearpage
\section*{SUPPLEMENTAL MATERIALS} \label{sec:supplemental}

\subsection*{Test Error for each model in each dataset}
\noindent The two models with the highest mean test Hinge Squared Error have been omitted for clearer visualization.
\foreach \x in {auto93, autohorse, autompg, autoprice, baskball, bodyfat, breasttumor, cholesterol, cleveland, cloud, cpu, echomonths, elusage, fishcatch, fruitfly, housing, lowbwt, lymphoma.mkatayama, lymphoma.tdh, machine.cpu, mbagrade, meta, pbc, pharynx, pollution, pwlinear, pyrim, sensory, servo, simulated.abs, simulated.linear, simulated.sin, sleep, stock, strike, triazines, veteran, vineyard, wisconsin} {
    \begin{figure}[!ht]
        \centering
        \includegraphics[width=0.8\textwidth]{figs.supplemental.\x_plot.pdf}
    \end{figure}
}

\clearpage
\subsection*{Ranking table of performance and consistency of each model}
\begin{table}[!ht]
\centering
\begin{tabular}{lrrrrrrr}
\toprule
 & constant & knn & linear & mmit & mmif & aft\_xgb & mlp \\
\midrule
auto93 & 7 & 6 & 3 & 5 & 1 & 2 & 4 \\
autohorse & 7 & 4 & 2 & 5 & 1 & 6 & 3 \\
autompg & 7 & 6 & 2 & 3 & 4 & 5 & 1 \\
autoprice & 7 & 4 & 6 & 2 & 1 & 5 & 3 \\
baskball & 6 & 2 & 5 & 3 & 1 & 7 & 4 \\
bodyfat & 7 & 6 & 2 & 4 & 3 & 5 & 1 \\
breasttumor & 6 & 1 & 4 & 5 & 2 & 7 & 3 \\
cholesterol & 3 & 5 & 4 & 6 & 1 & 7 & 2 \\
cleveland & 6 & 4 & 2 & 5 & 3 & 7 & 1 \\
cloud & 7 & 6 & 3 & 2 & 5 & 4 & 1 \\
cpu & 7 & 6 & 4 & 3 & 1 & 2 & 5 \\
echomonths & 7 & 4 & 2 & 5 & 3 & 6 & 1 \\
elusage & 7 & 2 & 1 & 3 & 4 & 5 & 6 \\
fishcatch & 7 & 6 & 4 & 5 & 3 & 2 & 1 \\
fruitfly & 1 & 5 & 2 & 6 & 3 & 7 & 4 \\
housing & 7 & 3 & 6 & 5 & 4 & 2 & 1 \\
lowbwt & 6 & 5 & 3 & 2 & 1 & 7 & 4 \\
lymphoma.mkatayama & 1 & 5 & 6 & 3 & 4 & 7 & 2 \\
lymphoma.tdh & 3 & 6 & 2 & 4 & 1 & 7 & 5 \\
machine.cpu & 7 & 6 & 5 & 3 & 2 & 1 & 4 \\
mbagrade & 3 & 6 & 5 & 1 & 2 & 4 & 7 \\
meta & 5 & 1 & 2 & 6 & 4 & 7 & 3 \\
pbc & 7 & 5 & 1 & 4 & 2 & 6 & 3 \\
pharynx & 7 & 6 & 2 & 1 & 3 & 4 & 5 \\
pollution & 4 & 6 & 3 & 2 & 1 & 7 & 5 \\
pwlinear & 7 & 6 & 5 & 2 & 3 & 4 & 1 \\
pyrim & 2 & 1 & 6 & 3 & 4 & 5 & 7 \\
sensory & 2 & 6 & 3 & 5 & 1 & 7 & 4 \\
servo & 7 & 6 & 4 & 2 & 5 & 1 & 3 \\
simulated.abs & 5 & 4 & 6 & 2 & 1 & 3 & 7 \\
simulated.linear & 7 & 6 & 1 & 3 & 2 & 5 & 4 \\
simulated.sin & 5 & 7 & 4 & 2 & 1 & 3 & 6 \\
sleep & 7 & 1 & 2 & 5 & 3 & 6 & 4 \\
stock & 7 & 2 & 5 & 3 & 1 & 6 & 4 \\
strike & 6 & 2 & 4 & 7 & 5 & 1 & 3 \\
triazines & 5 & 6 & 7 & 2 & 1 & 3 & 4 \\
veteran & 6 & 3 & 2 & 7 & 4 & 5 & 1 \\
vineyard & 7 & 4 & 3 & 6 & 5 & 1 & 2 \\
wisconsin & 6 & 5 & 4 & 2 & 1 & 7 & 3 \\
\bottomrule
\end{tabular}
\caption{The performance ranking of 7 models across 39 datasets.}
\end{table}

\begin{table}[!ht]
\centering
\begin{tabular}{lrrrrrrr}
\toprule
 & constant & knn & linear & mmit & mmif & aft\_xgb & mlp \\
\midrule
auto93 & 7 & 6 & 4 & 2 & 1 & 3 & 5 \\
autohorse & 7 & 4 & 2 & 5 & 1 & 6 & 3 \\
autompg & 7 & 5 & 1 & 4 & 2 & 6 & 3 \\
autoprice & 7 & 3 & 4 & 2 & 1 & 6 & 5 \\
baskball & 5 & 4 & 3 & 2 & 1 & 6 & 7 \\
bodyfat & 7 & 5 & 1 & 4 & 3 & 6 & 2 \\
breasttumor & 6 & 1 & 2 & 5 & 3 & 7 & 4 \\
cholesterol & 5 & 6 & 4 & 7 & 1 & 3 & 2 \\
cleveland & 6 & 5 & 1 & 4 & 3 & 7 & 2 \\
cloud & 7 & 6 & 3 & 2 & 5 & 4 & 1 \\
cpu & 7 & 4 & 5 & 2 & 1 & 3 & 6 \\
echomonths & 7 & 2 & 5 & 4 & 3 & 6 & 1 \\
elusage & 7 & 4 & 1 & 2 & 3 & 5 & 6 \\
fishcatch & 7 & 6 & 2 & 5 & 4 & 3 & 1 \\
fruitfly & 3 & 5 & 2 & 7 & 1 & 6 & 4 \\
housing & 7 & 3 & 4 & 5 & 2 & 6 & 1 \\
lowbwt & 6 & 4 & 3 & 1 & 2 & 7 & 5 \\
lymphoma.mkatayama & 1 & 6 & 5 & 2 & 3 & 7 & 4 \\
lymphoma.tdh & 3 & 6 & 4 & 2 & 1 & 7 & 5 \\
machine.cpu & 7 & 6 & 5 & 2 & 3 & 1 & 4 \\
mbagrade & 6 & 2 & 5 & 1 & 3 & 4 & 7 \\
meta & 6 & 1 & 2 & 5 & 4 & 7 & 3 \\
pbc & 6 & 2 & 3 & 5 & 4 & 7 & 1 \\
pharynx & 6 & 5 & 1 & 3 & 4 & 7 & 2 \\
pollution & 4 & 6 & 3 & 2 & 1 & 7 & 5 \\
pwlinear & 7 & 6 & 5 & 4 & 2 & 3 & 1 \\
pyrim & 1 & 3 & 6 & 4 & 5 & 2 & 7 \\
sensory & 1 & 6 & 3 & 4 & 2 & 7 & 5 \\
servo & 7 & 5 & 4 & 2 & 6 & 1 & 3 \\
simulated.abs & 6 & 7 & 5 & 2 & 1 & 3 & 4 \\
simulated.linear & 7 & 6 & 1 & 4 & 2 & 5 & 3 \\
simulated.sin & 4 & 7 & 6 & 2 & 1 & 3 & 5 \\
sleep & 7 & 4 & 2 & 5 & 1 & 6 & 3 \\
stock & 7 & 1 & 2 & 5 & 3 & 6 & 4 \\
strike & 5 & 7 & 6 & 3 & 2 & 1 & 4 \\
triazines & 1 & 5 & 3 & 6 & 2 & 7 & 4 \\
veteran & 7 & 1 & 4 & 3 & 6 & 5 & 2 \\
vineyard & 7 & 4 & 3 & 6 & 5 & 1 & 2 \\
wisconsin & 4 & 6 & 1 & 3 & 2 & 7 & 5 \\
\bottomrule
\end{tabular}
\caption{The consistency ranking of 7 models across 39 datasets.}
\end{table}

\end{document}